\documentclass[letterpaper, 10 pt, journal, twoside]{IEEEtran} 

\usepackage{microtype}
\usepackage{graphicx}
\usepackage{algorithmic}
\usepackage{algorithm}
\usepackage{array}
\usepackage{bm}
\usepackage[caption=false,font=footnotesize,labelfont=rm,textfont=rm]{subfig}
\usepackage{textcomp}
\usepackage{stfloats}
\usepackage{url}

\usepackage{verbatim}
\usepackage{setspace}
\usepackage{cite}
\usepackage{wrapfig,lipsum,booktabs}
\usepackage{amsmath,amssymb,stmaryrd,amsfonts} 
\usepackage[font=small]{caption, subcaption}
\usepackage{hyperref}
\usepackage{booktabs}
\usepackage{makecell}
\DeclareSubrefFormat{parens}{#1(#2)}

\begin{document}

\title{Kinematics-Aware Diffusion Policy with  \\ Consistent 3D Observation and Action Space \\ for Whole-Arm Robotic Manipulation}

\author{Kangchen Lv*, Mingrui Yu*, Yongyi Jia, Chenyu Zhang, and Xiang Li
\thanks{}}

\markboth{IEEE Robotics and Automation Letters. Submission Version.}
{Kinematics-Aware Diffusion Policy}


\maketitle

\begin{abstract}

Whole-body control of robotic manipulators with awareness of full-arm kinematics is crucial for many manipulation scenarios involving body collision avoidance or body-object interactions, which makes it insufficient to consider only the end-effector poses in policy learning.
The typical approach for whole-arm manipulation is to learn actions in the robot's joint space. 
However, the unalignment between the joint space and actual task space (i.e., 3D space) increases the complexity of policy learning, as generalization in task space requires the policy to intrinsically understand the non-linear arm kinematics, which is difficult to learn from limited demonstrations.
To address this issue, this letter proposes a kinematics-aware imitation learning framework with consistent task, observation, and action spaces, all represented in the same 3D space.
Specifically, we represent both robot states and actions using a set of 3D points on the arm body, naturally aligned with the 3D point cloud observations. 
This spatially consistent representation improves the policy's sample efficiency and spatial generalizability while enabling full-body control.
Built upon the diffusion policy, we further incorporate kinematics priors into the diffusion processes to guarantee the kinematic feasibility of output actions.
The joint angle commands are finally calculated through an optimization-based whole-body inverse kinematics solver for execution.
Simulation and real-world experimental results demonstrate higher success rates and stronger spatial generalizability of our approach compared to existing methods in body-aware manipulation policy learning. 

Project Website: \href{https://kinematics-aware-diffusion-policy.github.io}{\textit{kinematics-aware-diffusion-policy.github.io}}

\end{abstract}


\section{Introduction}



\IEEEPARstart{I}{mitation} learning, where an agent learns to mimic the expert demonstrations, is an efficient approach to acquire complex manipulation skills from limited data. Recently, diffusion-based visual-motor policies\cite{chi2024diffusion, Ze2024DP3, reuss2023goal} have shown many exciting results in imitation learning. 
Compared to traditional approaches, the remarkable abilities of diffusion models to learn multi-modal, high-dimensional action distributions are the key characteristics contributing to their success.

Due to the alignment between the action space and task space which simplifies the policy learning process, Cartesian-space end-effector pose representations are widely used in existing imitation learning methods.
However, for whole-arm robotic manipulation tasks, precise control over the full robot configuration is required, so imitating only the 6D end-effector pose trajectories is naturally insufficient. In many scenarios, such as operating in confined environments, avoiding collisions between the robot arm and surrounding obstacles is crucial.
Additionally, certain tasks require the robot to interact with objects using parts of its body rather than the end-effector, further necessitating the whole-body control.
Learning policies in joint space is a typical approach for whole-arm manipulation, which allows joint-level control of the entire configuration.
However, joint space is inherently unaligned with the task space where the manipulation is conducted, forcing the policy to implicitly understand the complex non-linear kinematics. Thus, it is hard to learn a generalizable joint-to-task mapping from limited demonstrations, restricting the sample efficiency and spatial generalizability of joint-space policies.





\begin{figure}[tbp]   
	\centering
	\includegraphics[width=8.5cm,scale=1.0]{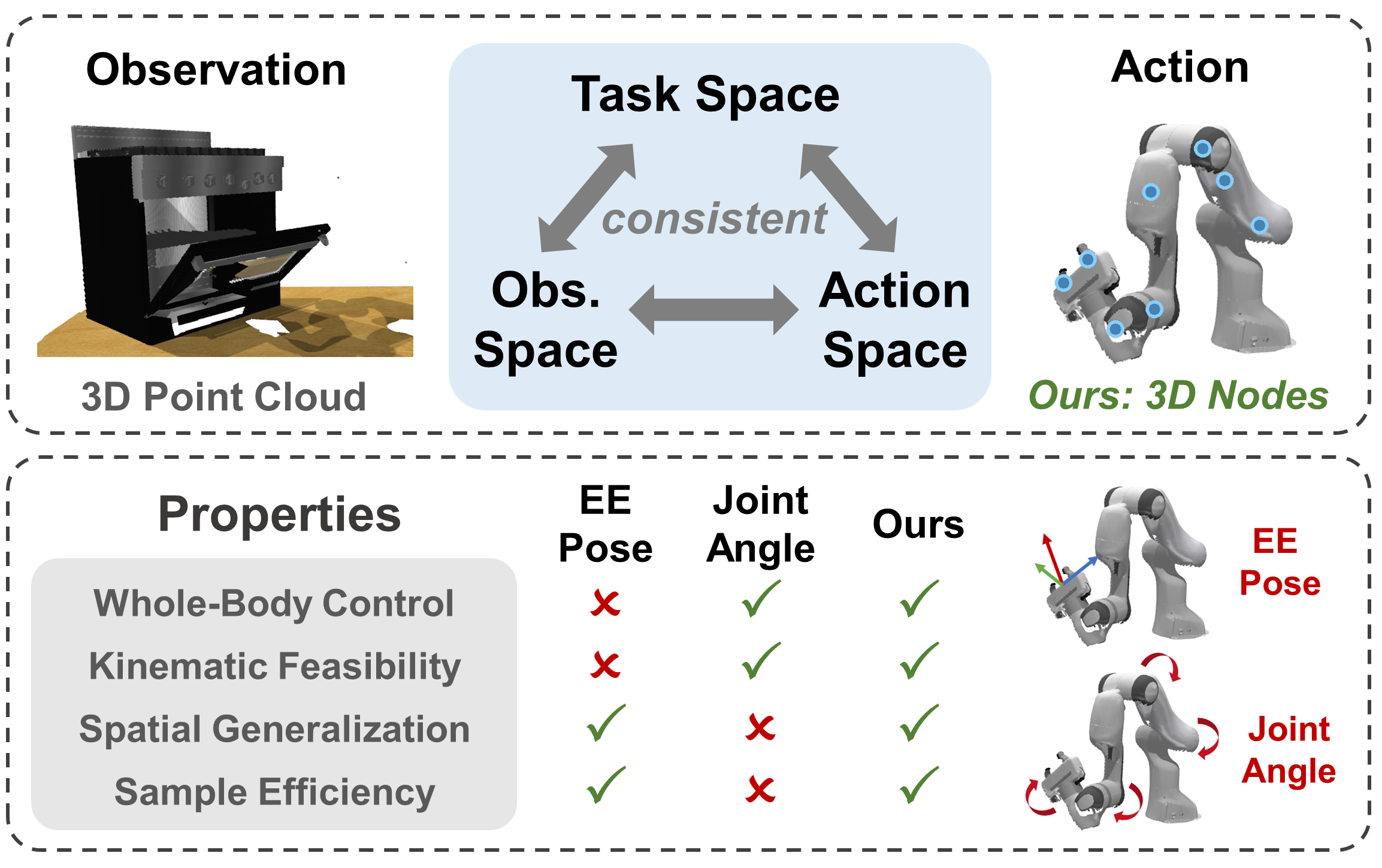}
	\caption{The proposed approach uses a set of 3D nodes on the arm body as both robot state and action representation for whole-arm manipulation, which is consistent with the 3D point cloud observation space and task space. Compared with using end-effector poses or joint angles, our method achieves higher spatial generalizability and sample efficiency while ensuring kinematic feasibility.}
	\label{fig:top}
\end{figure}


To improve the policy learning performance for whole-body manipulation, some previous works explore to combine Cartesian space and joint space via incorporating differentiable kinematics within the policy networks\cite{ma2024hierarchical,ganapathi2022implicit} or concatenating redundant joint states upon the end-effector poses\cite{mazzaglia2024redundancy}.
However, these methods still require the policy to predict joint-space actions, which cannot avoid the complexity brought by implicitly learning the non-linear kinematics.


In this paper, we propose \textbf{K}inematics-\textbf{A}ware \textbf{D}iffusion \textbf{P}olicy (\textbf{KADP}), with consistent task, observation, and action spaces. 
Instead of using joint angles, both robot states and actions are represented with a set of 3D nodes on the robot arm body, making it convenient for the policy to infer the spatial and geometric relationship between the robot configuration and point cloud observations in the same 3D space.
With such spatially consistent representation, the sample efficiency and spatial generalization of policy is improved while whole-body control is also enabled. 
To guarantee the kinematic feasibility of predicted 3D nodes, we further incorporate kinematic constraints into diffusion models.
For execution, the joint angle commands are finally computed through an optimization-based whole-body inverse kinematics solver.
In summary, the \textbf{\textit{kinematics awareness}} of the proposed policy learning approach attributes to the following three aspects:
\begin{enumerate}
 \item \textbf{Whole-Arm Control}: The proposed method enables manipulation over the entire robot configuration, overcoming the limitations of considering only Cartesian-space end-effector poses.
 \item \textbf{Consistent Task-Observation-Action Spaces}: The node representation is in the 3D space, consistent with the observation and task spaces, allowing the policy to directly infer the spatial relationship between the arm body, objects, and environments.
 \item \textbf{Kinematic Feasibility Guarantee}: By incorporating analytical joint-node mapping in both forward and reverse diffusion processes, our approach ensures that the generated node positions adhere to kinematic constraints.
\end{enumerate}

Across 8 simulation tasks on RLBench\cite{james2019rlbench} and 4 real-world tasks, we systematically evaluate the performance of the proposed approach, with comparison to several baseline methods using different action representations. KADP achieves higher success rate and stronger spatial generalizability, suggesting the effectiveness of utilizing such 3D node-based robot state and action representation in body-aware manipulation learning.




\section{Related Works}

\subsection{Diffusion Policies for Imitation Learning}

Diffusion models\cite{pmlr-v37-sohl-dickstein15, ho2020denoising} are a class of probabilistic generative models that learn to generate samples from the prior distribution, typically a Gaussian distribution, by an iterative denoising process. 
For visual imitation learning from demonstrations, Diffusion Policy\cite{chi2024diffusion} pioneers the generation of actions through a conditional diffusion model. 
This innovative formulation is able to effectively learn the multi-modal distribution of demonstration actions while ensuring training stability, which has also been employed as action decoding head in many large-scale generalist policy models such as Octo\cite{octo_2023}.
Subsequently, many follow-up works are introduced to improve the generalization ability, data efficiency and inference speed of diffusion policies.
DP3\cite{Ze2024DP3} and 3D Diffusion Actor\cite{ke2024d} enhance 3D scene representations by using 3D point cloud as observation space instead of RGB images, while some other works further leverage object-centric representations\cite{li2024language} or semantic fields\cite{wang2024gendp}. In this paper, we also adopt 3D point cloud as it has been proved to be more effective than images. 
Beyond vanilla diffusion models, BESO\cite{reuss2023goal} and PointFlowMatch\cite{chisari2024learning} build policies upon score-based diffusion model and flow matching perspective, respectively.
Besides, some works explore integrating several policies trained on heterogeneous data by composition\cite{wang2024poco, wang2024sparse} or accelerating diffusion policy with consistency distillation\cite{lu2024manicm, prasad2024consistency}.

\subsection{Kinematics-Aware Policy Learning}

For robotic manipulation, the selection of action spaces, such as Cartesian space, joint space, and torque space, will greatly influence the performance of various downstream tasks\cite{MartinMartin2019, Varin2019, Duan2021}.
Cartesian space, which controls the end-effector pose, is kinematics-unaware but aligns with the 3D Euclidean space in which the robot interacts with, whereas joint space provides complete low-level joint position control but increases the complexity of policy learning, in contrast\cite{aljalbout2024role}. 
Recently, some works are proposed to combine advantages of different action spaces for kinematics-aware policy learning.
Mazzaglia et al.\cite{mazzaglia2024redundancy} introduce a new family of action spaces for overactuated robot arms, which adds the joint position or angle of the redundant joint upon 6D end-effector pose.
IKP\cite{ganapathi2022implicit} links Cartesian space and joint space through forward kinematics to learn multi-action space policies.
Similarly but implemented in diffusion policy framework, HDP\cite{ma2024hierarchical} generates both end-effector pose and joint trajectories with two diffusion branches and finally refines joint positions from kinematics-unaware poses with differentiable kinematics.
Compared to previous works, we introduce a novel node-based representation in the 3D space consistent with the observation and task space, which avoids requiring the policy to implicitly learn the non-linear kinematics during predicting actions in joint space.



\subsection{Observation and Action Space Alignment}
Aligning the observation and action space, which can significantly simplify the observation-to-action mapping, has been shown as an effective way to improve sample efficiency and spatial generalization capability. In 2D image space, R\&D\cite{vosylius2024render} renders the gripper virutally in images to jointly represent RGB observations and actions, while Genima\cite{shridhar2024generative} draws joint actions as several colored spheres on RGB images and uses ACT\cite{zhao2023learning} as controller to translate visual targets to joint actions. 
Extending into 3D space, ActionFlow\cite{funk2024actionflow} introduces a new space consisting of object pose and feature sequences to represent both observation and action, but requires additional object pose estimators.
Some other methods utilize a simple same observation and action space such as intuitive 3D point clouds or voxels to avoid extra heavy computation cost for creating a new space.
For instance, C2F-ARM\cite{james2022coarse}, PerAct\cite{shridhar2023perceiver} and DNAct\cite{yan2024dnact} learn per-voxel features from discretized 3D observation and formulate the action prediction problem as a voxel classification task. Act3D\cite{gervet2023actd} and ChainedDiffuser\cite{xian2023chaineddiffuser} predict the next keyframe action by selecting a 3D point from uniformly-distributed point candidates, where observation and action lie in the same 3D space.
However, the methods above only consider the end-effector. In contrast, our proposed KADP enables whole-body control while preserving high spatial generalizability and sample efficiency afforded by the task-observation-action space alignment.

\section{Preliminaries}
\subsection{Problem Formulation} A standard imitation learning problem is considered here, where the goal is to learn an observation-to-action mapping $\pi:\mathcal{O}\xrightarrow{} \mathcal{A}$ from a set of expert demonstrations. Usually, the observation $O$ and action $A$ will both contain a few time steps, i.e. $O_t=\{o_{t-T_o+1},\cdots, o_{t-1}, o_t\}$ and $A_t=\{a_{t},a_{t+1},\cdots, a_{t+T_a-1}\}$, where $T_o$ is the length of observation history horizon and $T_a$ is the length of action prediction horizon. 
Given a demonstration dataset $D=\{(o_1,a_1,\cdots,o_{T_i},a_{T_i})\}_{i=1}^n$ consisting of $n$ trajectories with $\{T_i\}_{i=1}^n$ observation-action pairs, the imitation learning process is to train the visuomotor policy represented by a probability distribution $\pi(A|O)$ and then sample a robot action $A_t \sim \pi(A|O_t)$ from it during deployment.

\begin{figure*}[tbp]   
	\centering
	\includegraphics[width=17cm,scale=1.0]{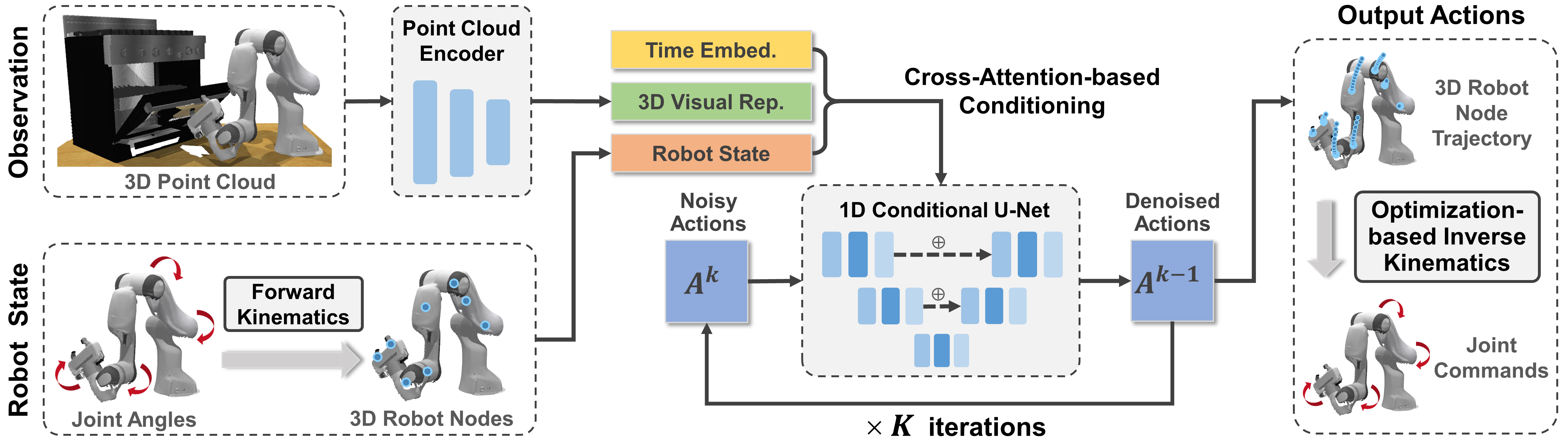}
	\caption{Overview of Kinematics-Aware Diffusion Policy (KADP). Taking the encoded 3D visual representations, the 3D robot nodes and time embeddings as input, diffusion model predicts the denoised 3D node trajectory iteratively. For execution, the joint angle commands are computed through an optimization-based whole-body inverse kinematics solver.}
	\label{fig:framework}
\end{figure*}

\subsection{Diffusion Policy for Action Generation} 

For the convenience of derivation in Sec.\ref{sec:method}, here we briefly introduce the diffusion policy\cite{chi2024diffusion} for action generation.
In the forward process, Gaussian noise is iteratively added to the action sample $A^0$ drawn from real distribution $q(A)$:
\begin{equation}
    q(A^k|A^{k-1}) := \mathcal{N}(A^k;\sqrt{1-\beta^k}A^{k-1},\beta^kI).
    \label{eq:1}
\end{equation}
Given the coefficients $\beta^1,\cdots, \beta^k$ determined by a noise scheduler and $\Bar{\alpha}^k=\prod _{i=1}^k(1-\beta^i)$, the noisy sample $A^k$ can be directly sampled from:
\begin{equation}
    q(A^k|A^0) := \mathcal{N}(A^k;\sqrt{\Bar{\alpha}^k}A^{0},(1-\Bar{\alpha}^k)I).
    \label{eq:2}
\end{equation}

Starting from an initial Gaussian noise $A^K\sim \mathcal{N}(0,I)$, the reverse process aims to construct the original noise-free data $A^0$ iteratively. Note that here the current observation $O$ is treated as the diffusion condition, so the parameterized model $p_{\theta}$ can be formulated as:
\begin{equation}
    p_{\theta}(A^{k-1}|A^k,O) := \mathcal{N}(A^{k-1};\mu_{\theta}(A^k,O,k),\Sigma_{\theta}(A^k,O,k)).
    \label{eq:3}
\end{equation}
At each diffusion step $k$, a denoising network $\epsilon_{\theta}$ parameterized by $\theta$ is trained to predict the noise component of $A^k$. The iterative denoising process is
\begin{equation}\label{eq:dp_a}
    A^{k-1} =\alpha_k(A^k-\gamma_k\epsilon_\theta(A^k,O,k)) + \sigma_k\mathcal{N}(0,I).
\end{equation}

Based on Eq.\ref{eq:2} and Eq.\ref{eq:3}, the model can be trained by maximizing the evidence lower bound (ELBO):
\begin{equation}
    \mathbb{E}_{A^0}\,{\rm log}\,p_{\theta}(A^0) \geq \mathbb{E}_{q(A^{1:K}|A^0)}[\,{\rm log} \frac{p_{\theta}(A^{0:K})}{q(A^{1:K}|A^0)}].
\end{equation}
 During training, we randomly sample a data $A^0$ and add noise $\epsilon^k$ over $k$ steps through the forward process.
 The training objective can be derived to minimize the difference between the added noise and the network $\epsilon_\theta$ prediction:
\begin{equation}
    \mathcal{L}={\rm MSELoss}(\epsilon^k, \epsilon_\theta(A^k,O,k)).
\end{equation}

\section{Method}\label{sec:method}
\subsection{3D Node-Based Robot State and Action Representation}

We introduce a set of 3D nodes to represent the robot configuration, denoted as ${A}_{\rm node} = \{(x_0,y_0,z_0),\cdots,(x_m,y_m,z_m)\}$, where $(x_i,y_i,z_i)$ corresponds to the coordinates of the $i^{\rm th}$ joint and $m$ is the number of selected nodes.
This novel node-based representation is defined in the 3D Euclidean space, consistent with the point cloud observation space and task space, 
allowing the denoising network $\epsilon_{\theta}$ to learn  within the same 3D space and thus improving its sample efficiency and spatial generalizability.
The principle of node selection is to fully describe the robot configuration with the minimal number of nodes. For the 7-DoF Franka Emika Panda robot arm, we manually choose 8 nodes as shown in the bottom left of Fig.\ref{fig:framework}.
The first 6 nodes are located on the robot arm from the $1^{\rm st}$ joint (the base) to the $6^{\rm th}$ joint, ensuring the state of each joint is reflected by the 3D position of the corresponding node. 
We further place two extra nodes on the left/right gripper fingers to represent both the states of the $7^{\rm th}$ joint and gripper. Notably, we also place an additional binary value indicating the gripper's open/close action in the node-based representation as discrete control of the gripper is empirically found to be more effective. For writing brevity, we will omit the straight-forward implementation of it in the following sections.


Note that the entire robot joint configuration is uniquely determined given feasible 3D node positions, which enables whole-body control in contrast with end-effector-based policies.
In addition, the space alignment between 3D point cloud observations and node-based states/actions offers higher sample efficiency and stronger spatial generalizability compared to joint-space policies. 
For instance, when the positions of manipulated objects change, the node-based policy can straightforwardly interpret the spatial relationship between new point cloud observations and 3D node positions. In contrast, reflecting task-space object variations in joint space is non-linear and complex, making joint-space policy learning more difficult.

In the diffusion policy framework, we can seamlessly take such 3D nodes as both the state and action representation for conditional action generation. For robot state, the corresponding node positions can be easily computed from joint angles via forward kinematics, defined by the mapping ${\rm F}_{\rm fk}(\cdot):\mathbb{R}^{n}\rightarrow\mathbb{R}^{3\times m}$. For execution, the joint angle commands are required to be transferred from the predicted 3D node trajectory, denoted as ${\rm F}_{\rm ik}(\cdot):\mathbb{R}^{3\times m}\rightarrow\mathbb{R}^{n}$. We achieve this through an optimization-based inverse kinematics solver. Given the predicted 3D node positions ${A}_{\rm node}$, the optimization for joint angle commands ${A}_{\rm joint}$ is formulated as:
\begin{equation}\label{eq:opt}
\begin{aligned}
    \mathop{\min}_{{A}_{\rm joint}}\; &||\,\Lambda\cdot \left( {\rm F}_{\rm fk}({A}_{\rm joint})-{A}_{\rm node}\right)\,||_2^2     \\
     {\rm s.t.}\;& \Theta_{\rm min} \leq {A}_{\rm joint}\leq \Theta_{\rm max},
\end{aligned}
\end{equation}
where $\Theta_{\rm min}$ and $ \Theta_{\rm max}$ are the joint limits, and $\Lambda={\rm diag} (\lambda_1,\cdots,\lambda_m)$ is a diagonal weight matrix.



\subsection{Diffusion Model with Kinematic Constraints}\label{sec:kc}
Inherently, the 3D node representation is redundant with respect to the actuated DoFs of the arm. Thus, the original diffusion policy cannot guarantee that the generated 3D node positions correspond to a valid robot configuration, where the potential kinematic infeasibility will lead to inaccurate optimized joint commands and affect the manipulation performance.
Consequently, We further incorporate kinematic constraints directly into diffusion models, which ensures that the node positions are kinematic feasible throughout the training and inference process.

Inspired by related works\cite{chisari2024learning, jiang2023se} exploring variations of the diffusion model on $SO(3)$ or $SE(3)$ manifold, we define the distance of two node representations within the transferred compact joint space, rather than the original 3D Euclidean space. 
The interpolation operation between the start nodes $A^0$ and target nodes $A^1$ is then expressed as $A^t= {\rm F}_{\rm fk}(\,t\, {\rm F}_{\rm ik}(A^0)+(1-t)\, {\rm F}_{\rm ik}(A^1))$.
Similarly, the noise perturbation of node representation is also defined on joint space and then transferred to nodes, so that the forward process can be denoted as:
\begin{equation}
    A^k= {\rm F}_{\rm fk}(\sqrt{\Bar{\alpha}^k} \,{\rm F}_{\rm ik}(A^0) +\sqrt{1-\Bar{\alpha}^k}\,\epsilon),
    \label{eq:forward}
\end{equation}
where the standard Gaussian noise $\epsilon \sim \mathcal{N}(0,I)$. 

Following DDPM\cite{ho2020denoising}, the posterior distribution of  can be derived using Bayes’ rule as:
\begin{equation}\label{eq:reparm}
     q({\rm F}_{\rm ik}(A^{k-1})|A^{k},A^{0}) := \mathcal{N}({\rm F}_{\rm ik}(A^{k-1}); \Tilde{\mu}^k(A^0, A^k), \Tilde{\beta}^kI),
\end{equation}
where $ \Tilde{\mu}^k(A^0, A^k)=\frac{\sqrt{\Bar{\alpha}^{k-1}}\beta^k}{1-\Bar{\alpha}^k} {\rm F}_{\rm ik}(A^0) +\frac{\sqrt{\alpha^{k}}(1-\Bar{\alpha}^{k-1})}{1-\Bar{\alpha}^k}{\rm F}_{\rm ik}(A^k)$
 and $\Tilde{\beta}^k=\frac{(1-\Bar{\alpha}^{k-1})\beta^k}{1-\Bar{\alpha}^{k}}$. We also follow the DP3\cite{Ze2024DP3} to use sample prediction instead of epsilon prediction for better high-dimensional action generation, so the objective for training network $\mu_\theta$ is:
 \begin{equation}\label{eq:obj}
    \mathcal{L}={\rm MSELoss}(A^0, \,\mu_\theta(A^k,O,k)).
\end{equation}

\begin{algorithm}[tb]
\caption{Training Procedure of KADP}
\label{alg:A}\setstretch{1.1}
\begin{algorithmic}
\REPEAT
\STATE {$(O,A^0)\sim D$ \hfill $\triangleright$ sample dataset} 
\STATE {$k \leftarrow {\rm Randint}(0,K)$\hfill $\triangleright$ sample diffusion step}
 \STATE {$\epsilon\sim \mathcal{N}(0,I)$\hfill $\triangleright$ sample noise} 
 \STATE {$ A^k= {\rm F}_{\rm fk}(\sqrt{\Bar{\alpha}^k} \,{\rm F}_{\rm ik\_mlp}(A^0) +\sqrt{1-\Bar{\alpha}^k}\,\epsilon)$\hfill $\triangleright$ Eq.\ref{eq:forward}} 
\STATE $ {L}={\rm MSELoss}(A^0, \,\mu_\theta(A^k,O,k))$\hfill $\triangleright$ Eq.\ref{eq:obj}
\STATE $ \theta=\theta-\alpha\nabla_{\theta}\,L$\hfill $\triangleright$ update network params
\UNTIL{$\mu_\theta$ converged}
\end{algorithmic}
\end{algorithm}

\begin{algorithm}[tb]
\caption{Sampling Procedure of KADP}
\label{alg:B}\setstretch{1.1}
\begin{algorithmic}
 \STATE {$A^K\sim \mathcal{N}(0,I)$ \hfill $\triangleright$ sample starting point} 
 \FOR{$k = K$ to $1$}
 \STATE {$z\sim \mathcal{N}(0,I)$ \hfill $\triangleright$ sample noise}
 \STATE {$\hat{A}^0=\mu_\theta(A^k,O,k)$ \hfill $\triangleright$ network prediction}\vspace{1ex}
 \STATE {$\Tilde{\mu}^k=\frac{\sqrt{\Bar{\alpha}^{k-1}}\beta^k}{1-\Bar{\alpha}^k} {\rm F}_{\rm ik\_opt}(\hat{A}^0) +\frac{\sqrt{\alpha^{k}}(1-\Bar{\alpha}^{k-1})}{1-\Bar{\alpha}^k}{\rm F}_{\rm ik\_opt}(A^k)$ }\vspace{0.8ex}
 \STATE {$A^{k-1}= {\rm F}_{\rm fk}\big(\Tilde{\mu}^k + \sqrt{\Tilde{\beta}^k}\,z\big)$ \hfill $\triangleright$ Eq.\ref{eq:sampling}} 
 \ENDFOR
 \RETURN {$A^0$}
\end{algorithmic}
\end{algorithm}

To make the network trainable, the differentiability of the mappings ${\rm F_{fk}}$ and ${\rm F_{ik}}$ in Eq.\ref{eq:obj} is required. For the joint-to-node mapping ${\rm F_{fk}}$, differentiable forward kinematics with a predefined robot URDF model can provide gradients. However, the node-to-joint mapping implemented by optimization-based inverse kinematics solver, denoted as ${\rm F_{ik\_opt}}$, is non-differentiable, preventing gradients from passing through. To address this, we pretrain a 3-layer MLP, denoted as ${\rm F_{ik\_mlp}}$, to fit this inverse kinematics mapping and then freeze it during policy model training. Compared with the optimization-based ${\rm F_{ik\_opt}}$, the MLP-based ${\rm F_{ik\_mlp}}$ is differentiable but less accurate. Thus, ${\rm F_{ik\_mlp}}$ is only used to offer approximate gradients during training and ${\rm F_{ik\_opt}}$ is employed for accurate node-to-joint mapping during inference.

Starting from a noise $A^K\sim \mathcal{N}(0,I)$, action generation, which is modeled as the iterative denoising process, also follows the Diffusion Policy framework. The predicted original sample $\hat{A}^0=\mu_\theta(A^k,O,k)$ is used to compute the mean value of the distribution of $A^{k-1}$ in Eq.\ref{eq:reparm}. The sampling process can be written as:
\begin{equation}\label{eq:sampling}
     A^{k-1}= {\rm F}_{\rm fk}\big(\,\Tilde{\mu}^k(\hat{A}^0, A^k) + \sqrt{\Tilde{\beta}^k}\,z\big),
\end{equation}
where $z\sim \mathcal{N}(0,I)$ represents the random Gaussian noise.
In practice, DDIM\cite{song2021denoising} is commonly utilized to accelerate the generation process with non-Markovian diffusion processes.

The overview of training and sampling procedure is shown in Alg.\ref{alg:A} and Alg.\ref{alg:B}. As aforementioned, the MLP-based ${\rm F_{ik\_mlp}}$ and optimization-based ${\rm F_{ik\_opt}}$ are used during training and sampling, respectively. Note that although the node representation is initially transferred to joint space and later recovered, the space alignment between observation and action is still maintained throughout the action generation process as both input and output of noise prediction network are within the consistent 3D space.

\subsection{Model Architecture}
The 3D point cloud observation is first encoded into visual representations with an MLP-based encoder, which has been shown to be simple but effective in DP3\cite{Ze2024DP3}, and then concatenated with the robot proprioception state to form the conditional information. We choose cross attention layers instead of the classic FiLM layers\cite{perez2018film} for conditioning.
The noisy action $A^k$, observation embedding and the positional embedding of the diffusion step $k$ are then passed into the 1D U-Net, which output the predicted original sample $\hat{A}^0$. Then, the one-step denoised action $A^{k-1}$ can be computed.

\begin{figure*}[tbp]   
	\centering
	\includegraphics[width=18cm,scale=1.0]{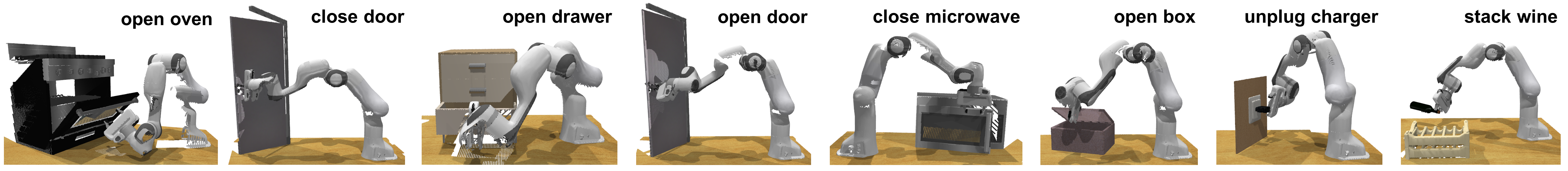}
	\caption{Visualization of 8 RLBench simulation tasks.}
	\label{fig:overview_tasks}
\end{figure*}

\begin{table*}
\centering
\caption{Performance of our proposed KADP and the baselines (DP3-EE, DP3-Joint, and DP3-ERJ) on 8 RLBench simulation tasks.}
\begin{tabular}{l|p{1.3cm}<{\centering} p{1.3cm}<{\centering} p{1.3cm}<{\centering} p{1.3cm}<{\centering} p{1.3cm}<{\centering} p{1.5cm}<{\centering }p{1.3cm}<{\centering} p{1.45cm}<{\centering}|c}
\toprule
\small Method & \makecell[c]{\small$\mathtt{open}$\\ \small$\mathtt{oven}$}&
\makecell[c]{\small$\mathtt{open}$\\ \small$\mathtt{drawer}$}&
\makecell[c]{\small$\mathtt{open}$\\ \small$\mathtt{box}$}&
\makecell[c]{\small$\mathtt{open}$\\ \small$\mathtt{door}$}&
\makecell[c]{\small$\mathtt{close}$\\ \small$\mathtt{door}$}&
\makecell[c]{\small$\mathtt{close}$\\ \small$\mathtt{microwave}$}&
\makecell[c]{\small$\mathtt{unplug}$\\ \small$\mathtt{charger}$}&
\makecell[c]{\small$\mathtt{stack}$\\ \small$\mathtt{wine}$}&
\small Average\\ 
\midrule
\small DP3-EE&
\small24.3{\scriptsize\ $\pm$2.1}&
\small75.7{\scriptsize\ $\pm$1.5}&
\small70.7{\scriptsize\ $\pm$2.5}&
\small33.3{\scriptsize\ $\pm$2.5}&
\small6.0{\scriptsize\ $\pm$1.0}&
\small6.7{\scriptsize\ $\pm$0.6}&
\small\textbf{36.7}{\scriptsize\ $\pm$5.7}&
\small68.0{\scriptsize\ $\pm$2.0}&
\small40.2{\scriptsize\ $\pm$0.5}\\ 

\small DP3-Joint&
\small31.3{\scriptsize\ $\pm$4.0}&
\small23.3{\scriptsize\ $\pm$4.2}&
\small\textbf{77.7}{\scriptsize\ $\pm$5.0}&
\small28.0{\scriptsize\ $\pm$2.0}&
\small18.7{\scriptsize\ $\pm$2.5}&
\small79.7{\scriptsize\ $\pm$4.0}&
\small15.3{\scriptsize\ $\pm$1.5}&
\small68.3{\scriptsize\ $\pm$2.1}&
\small42.8{\scriptsize\ $\pm$1.3}\\

\small DP3-ERJ&
\small23.7{\scriptsize\ $\pm$3.2}&
\small55.7{\scriptsize\ $\pm$3.2}&
\small68.0{\scriptsize\ $\pm$2.6}&
\small37.0{\scriptsize\ $\pm$3.6}&
\small15.0{\scriptsize\ $\pm$2.0}&
\small52.3{\scriptsize\ $\pm$0.6}&
\small30.7{\scriptsize\ $\pm$2.5}&
\small\textbf{75.3}{\scriptsize\ $\pm$3.1}&
\small44.7{\scriptsize\ $\pm$0.9}\\

\midrule

\small KADP (Ours)&
\small\textbf{51.3}{\scriptsize\ $\pm$2.1}&
\small\textbf{92.7}{\scriptsize\ $\pm$2.3}&
\small76.3{\scriptsize\ $\pm$1.5}&
\small\textbf{55.0}{\scriptsize\ $\pm$2.0}&
\small\textbf{50.0}{\scriptsize\ $\pm$2.6}&
\small\textbf{87.3}{\scriptsize\ $\pm$2,9}&
\small31.3{\scriptsize\ $\pm$4.5}&
\small70.7{\scriptsize\ $\pm$3.2}&
\small\textbf{64.3}{\scriptsize\ $\pm$1.3}\\
\bottomrule
\end{tabular}\label{tab:1}
\end{table*}

\section{Simulation Evaluation}




\subsection{Evaluation Settings}\label{sec:eval}

From the popular robot learning benchmark RLBench\cite{james2019rlbench}, we pick 8 challenging tasks for evaluation. Almost all the selected tasks are difficult to execute with only end-effector control, while whole-body control contributes a lot to successful manipulation.
The standard expert demonstration collection interface in RLBench is utilized to collect 20 trajectories for each task. 
The resolution of RGB-D images captured by five multi-view cameras is 128 $\times$ 128, from which the object region is segmented and projected to 3D space as the point cloud observation. For batch training, we downsample the point cloud to 1024 points with Farthest Sampling Point algorithm.

All the policy models are trained for 3000 epochs on each task with AdamW optimizer, where the learning rate is 1e-4 and the weight decay is 1e-6. Other hyper-parameters include the observation horizon $T_o=2$, action horizon $T_a=8$ and the execution horizon $T_e=4$. During training, 100 diffusion denoising steps are performed while 10 denoising steps with DDIM noise scheduler are used for inference.


\begin{figure}[!t]
\centering
\subfloat[Kinematic Constraints in DP]{\includegraphics[height=0.32\linewidth]{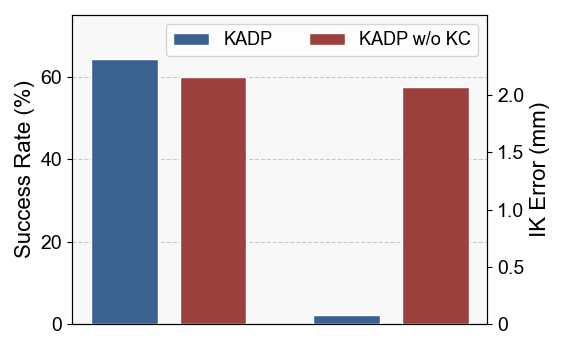}\label{fig:ablation_kc}}
\hfill
\subfloat[The Number of Nodes]{\includegraphics[height=0.32\linewidth]{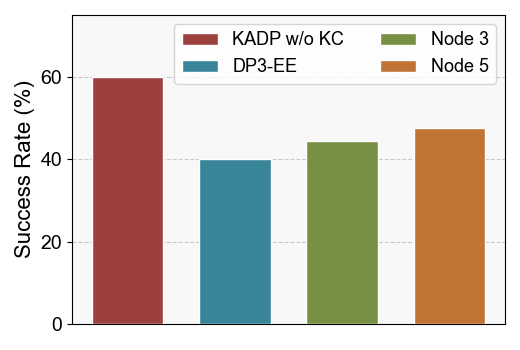}\label{fig:ablation_node_num}}
\caption{Ablation on the kinematic constraints in DP and the number of nodes. IK Error refers to the average per-node inverse kinematics optimization error when solving joint commands. \textit{KADP \textit{w/o} KC}: remove the kinematic constraints in DP. \textit{Node-3\,/\,Node-5}: replace the full 8 nodes with fewer nodes.}
\label{fig:ablation}
\end{figure}

\subsection{Comparison with Baselines}
We compare KADP against the following baselines: 1) 3D Diffusion Policy using Cartesian-Space End-Effector Poses (DP3-EE): Predicting a sequence of end-effector poses and computing correspondent joint commands via inverse kinematics solvers; 2) 3D Diffusion Policy using Joint Space (DP3-Joint): Predicting a sequence of joint angles; 3) 3D Diffusion Policy using ERJ Space (DP3-ERJ): Using the ERJ space\cite{mazzaglia2024redundancy} which concatenates the 6D end-effector pose and the joint positions corresponding to redundant joints in robot arms. For fair comparison, all the settings described in Sec.\ref{sec:eval} are kept identical, except for the robot state and action representations. The average success rate and the standard deviation with 3 individual evaluation runs on 100 episodes are reported.

As shown in Table \ref{tab:1}, although the performance of baseline methods shows variability among tasks, KADP consistently achieves the best or second-best performance on all tasks with an overall average success rate of 64.3\%, showcasing an improvement of nearly 20\% over the baseline methods. These experimental results demonstrate that KADP, benefiting from whole-arm control and the alignment between task, observation, and action spaces, is capable of considering whole-arm motion while maintaining high sample efficiency and strong spatial generalization ability. 
Among these three baselines, DP3-EE performs worst, highlighting the limitations of considering only the end-effector pose. DP3-Joint shows comparable low performance, which demonstrates its lower sample efficiency and spatial generalizability brought by the inconsistent spaces. 
With access to the redundant joint, DP3-ERJ, also enabling full-body control, performing slightly better but its partial alignment between the observation and task space and incomplete kinematics awareness constrains its effectiveness compared to our proposed approach.

In our experiments, several typical failure modes of the baseline methods are observed. For instance, DP3-EE reaches only 6.7\% on the $\mathtt{close}$ $\mathtt{microwave}$ task, where most failure cases stem from the generated kinematically infeasible end-effector poses. The ERJ action space, which introduces additional constraints for the redundant joint, can  significantly reduce the IK errors and boost the performance of end-effector-based policies with a 46.3\% improvement on this task. On the $\mathtt{open}$ $\mathtt{oven}$ task, controlling only the end-effector pose by DP3-EE frequently causes the arm body to collide with the oven door during lifting stage. In contrast, DP3-Joint enables smoother control of individual joints, but suffers from inaccurate task-space generalization, leading to more frequent failure grasping of the oven's thin handle.




\begin{figure*}[!t]
\centering
\subfloat[Pick up Cube]{\includegraphics[width=0.492\linewidth]{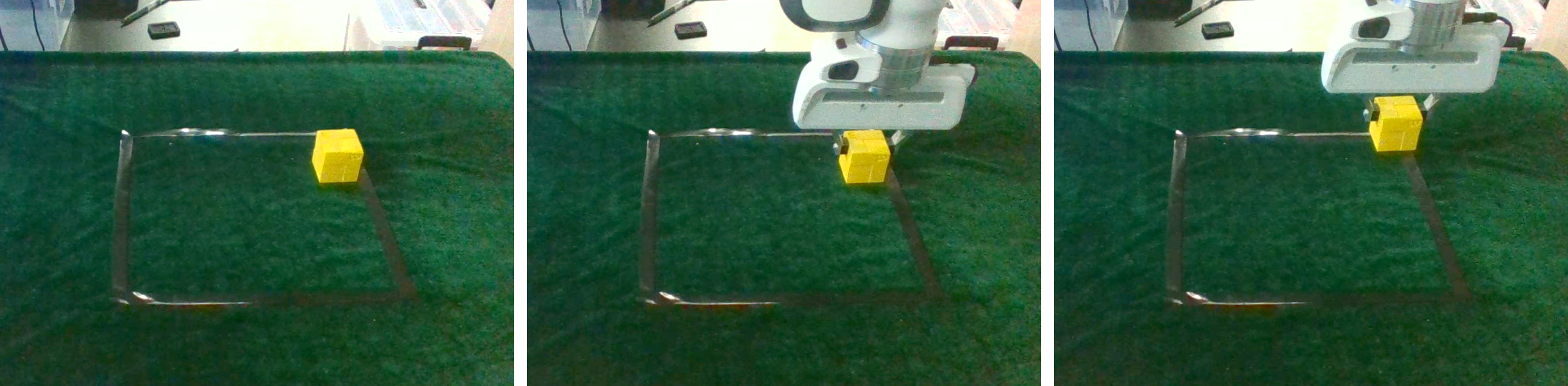}\label{fig:pick}}
\hfill
\subfloat[Open Door]{\includegraphics[width=0.492\linewidth]{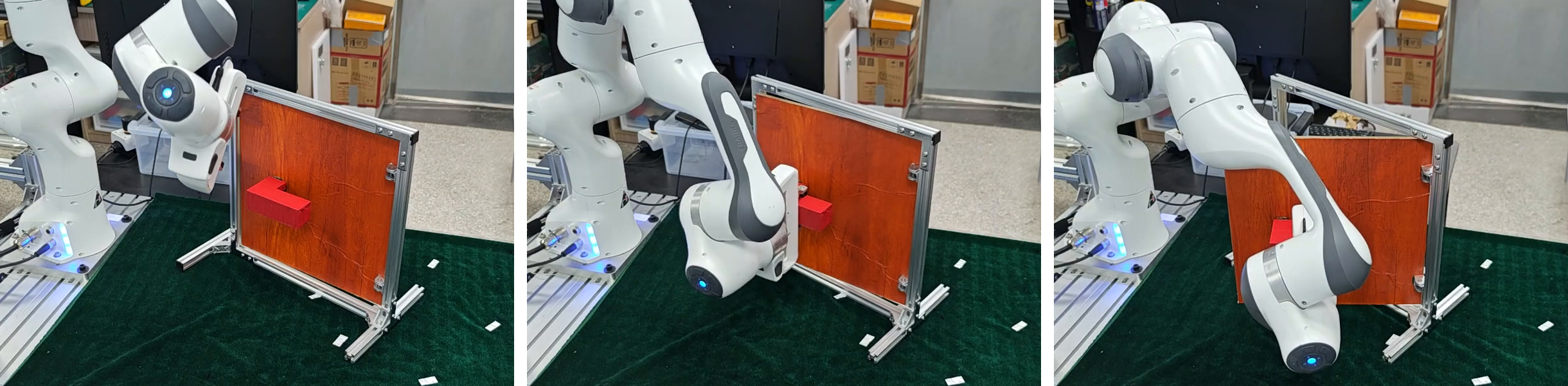}}
\hfill\\
\subfloat[Put Cube in Cabinet]{\includegraphics[width=0.492\linewidth]{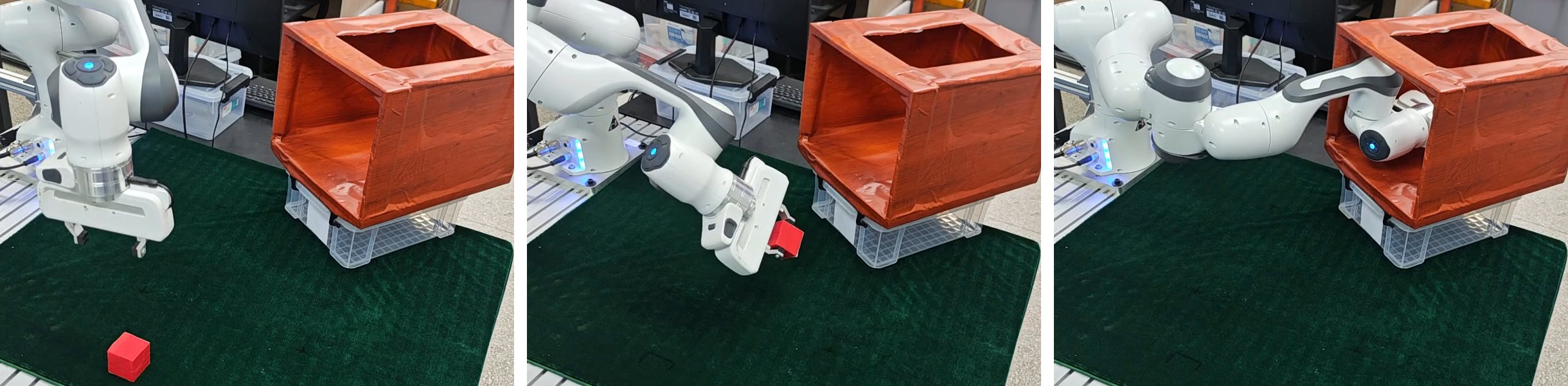}}
\hfill
\subfloat[Push Button Elbow]{\includegraphics[width=0.492\linewidth]{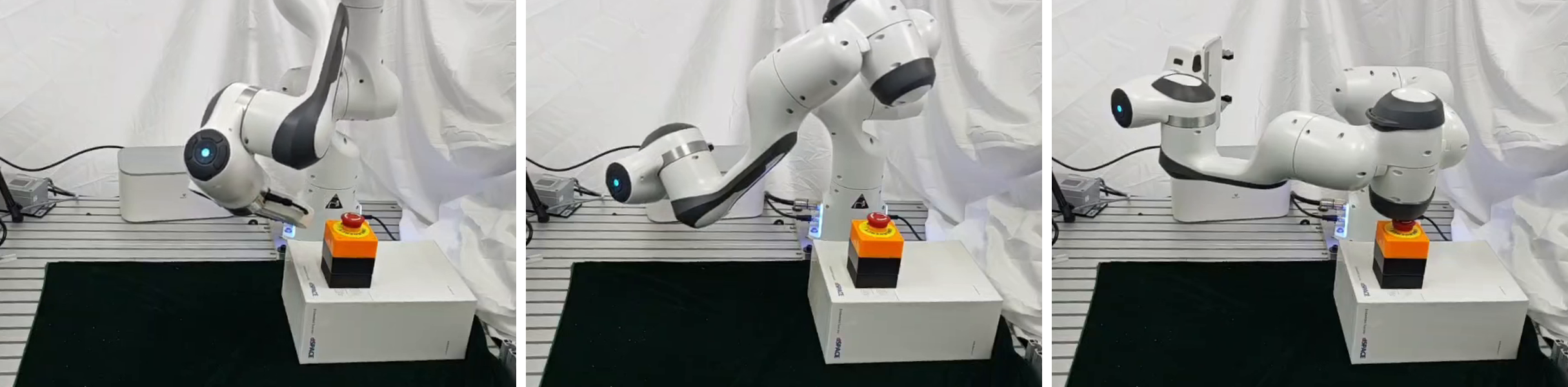}}
\caption{Overview of the 4 real-world tasks, where the manipulation is achieved by the proposed KADP.}
\label{fig:realworld}
\end{figure*}

\subsection{Ablation Studies}

\textbf{Kinematic Constraints in DP}: 
Firstly, we conduct an ablation study on the effect of kinematics-aware diffusion process proposed in Sec.\ref{sec:kc}. 
To assess the kinematic feasibility of the generated node positions, we calculate the average per-node distance (i.e., IK error) between the diffusion policy's predicted 3D nodes and those corresponding to joint angles obtained by the optimization-based inverse kinematics solver.
When taking the 3D nodes directly as the state and action representation in diffusion policy, the predicted actions cannot be theoretically guaranteed to be kinematically feasible. 
As shown in Fig.\ref{fig:ablation_kc}, the predicted node positions remain approximately kinematically feasible in practice with an average IK error of 2.1mm, even in the absence of explicit constraints.
In contrast, the full KADP framework, which incorporates kinematic constraints into the diffusion process, reduces the IK error to nearly zero and slightly improve the task success rate by 4.3\%.

 \textbf{The Number of Nodes}:
We also perform an ablation study on the number of nodes, where two reduced node sets, referred to as Node-3 and Node-5, are considered here.
In addition to the 2 gripper nodes, Node-3 includes another node on the $6^{\rm th}$ joint, while Node-5 includes three nodes on the $1^{\rm st}, 4^{\rm th}$, and $6^{\rm th}$ joints. 
Since the full robot joint configuration cannot be uniquely determined with only 3 nodes, MLP-based inverse kinematics is not applicable in this case. Therefore, we compare KADP without kinematic constraints to these two ablated settings. 
As provided in Fig.\ref{fig:ablation_node_num}, Node-3 achieves an average success rate of 44.5\%, comparable to DP3-EE. 
This is expected, as Node-3 can only represent the gripper’s translation and orientation, which is roughly equivalent to Cartesian-space end-effector pose.
Node-5, on the other hand, achieves a higher performance of 47.6\%, approaching the performance of using 8 nodes more closely. This result further validates the advantage of representing the full robot configuration in 3D space for effective policy learning.

\section{Real-world Experiments}


\subsection{Environment Setup}
A 7-DoF Franka Emika Panda robot arm is employed as the real-world platform, equipped with a fixed front-view RealSense D435 camera to capture point cloud observations.
All hyper-parameters are kept consistent with those used in the simulation studies, except for the action prediction horizon $T_a = 4$. 
Snapshots of 4 designed real-world tasks, which evaluate different capabilities of our method, are shown in Fig.\ref{fig:realworld}.
For all tasks except $\mathtt{pick\ up \ cube}$, we collect 10 expert demonstrations for training and perform 10 evaluation trials per task, with randomized object poses every time. 
The average inference time cost for the denoising process is 0.13s on an NVIDIA RTX 3090 GPU, while the optimization-based IK solver takes an average of 2.8ms per call. Thus, we run the policy at 5Hz and control the robot at 10Hz by executing the first two actions in the predictions.
The 4 tasks are as follows:

\textbf{Pick up Cube}: The robot only needs to grasp the object and lift it, which is designed to specifically analyze the spatial generalizability and sample efficiency of KADP. Since accurately controlling the end-effector pose is sufficient, DP3-EE is expected to perform well due to the alignment between observation and action space.

 
\textbf{Open Door}: The robot should first grasp the handle and then follow a circular trajectory to open the door. Due to the narrow width of the handle, even small grasping positional error from the handle’s center will cause the gripper to lose contact with it in the subsequent motion. Controlling only the end-effector pose is also sufficient but this task is obviously more challenging compared to the $\mathtt{pick \ up \ cube}$ above.

\textbf{Put Cube in Cabinet}: The robot should first grasp a cube and then put it into a deep and narrow cabinet, which is designed to evaluate whole-body collision avoidance performance. The primary difficulties arise from two factors: 1) The robot must reach near the cabinet's deepest position, which requires the entire robot to remain nearly horizontal to avoid collision with the top surface; 2) The cabinet is only 4cm wider than the gripper, making the successful insertion highly sensitive to even slight positional inaccuracies.


\textbf{Push Button Elbow}: The robot is required to press a button using its elbow instead of the gripper, making it meaningless to control only the end-effector pose. Learning directly from joint space is expected to yield good performance as only the angles of the first 3 joints change during the manipulation process.

\subsection{Experimental Results and Comparison with Baselines}

\begin{table*}[tb]
\caption{Performance of the proposed KADP and baselines on 4 real-world tasks.}
\centering
\begin{tabular}{l|p{2.5cm}<{\centering} p{2.5cm}<{\centering} p{2.25cm}<{\centering} p{2.5cm}<{\centering} p{2.5cm}<{\centering}} 
\toprule
Method & \makecell[c]{$\mathtt{pick\; up\;cube}$\\(5 demos)} & \makecell[c]{$\mathtt{pick\; up\;cube}$ \\ (13 demos)} &  \makecell[c]{$\mathtt{open\;door}$} &  \makecell[c]{$\mathtt{put\;cube}$ \\ $\mathtt{in\; cabinet}$} &  \makecell[c]{$\mathtt{push\;button}$\\ $\mathtt{elbow}$} \\ 
\midrule
DP3-EE & 13\;\!/\;\!25 & 19\;\!/\;\!25 &  \textbf{8\;\!/\;\!10} & 1\;\!/\;\!10 & 0\;\!/\;\!10\\
DP3-Joint & 6\;\!/\;\!25& 10\;\!/\;\!25 & 6\;\!/\;\!10 & 7\;\!/\;\!10 & \textbf{10\;\!/\;\!10}\\
KADP (Ours) & \textbf{15\;\!/\;\!25} & \textbf{22\;\!/\;\!25} &  \textbf{8\;\!/\;\!10} & \textbf{9\;\!/\;\!10} & \textbf{10\;\!/\;\!10}\\
\bottomrule
\end{tabular}\label{tab:2}
\end{table*}

\begin{figure}[tb]
\centering
\subfloat[Pick up Cube: 5 demonstrations]{\includegraphics[width=1.0\linewidth]{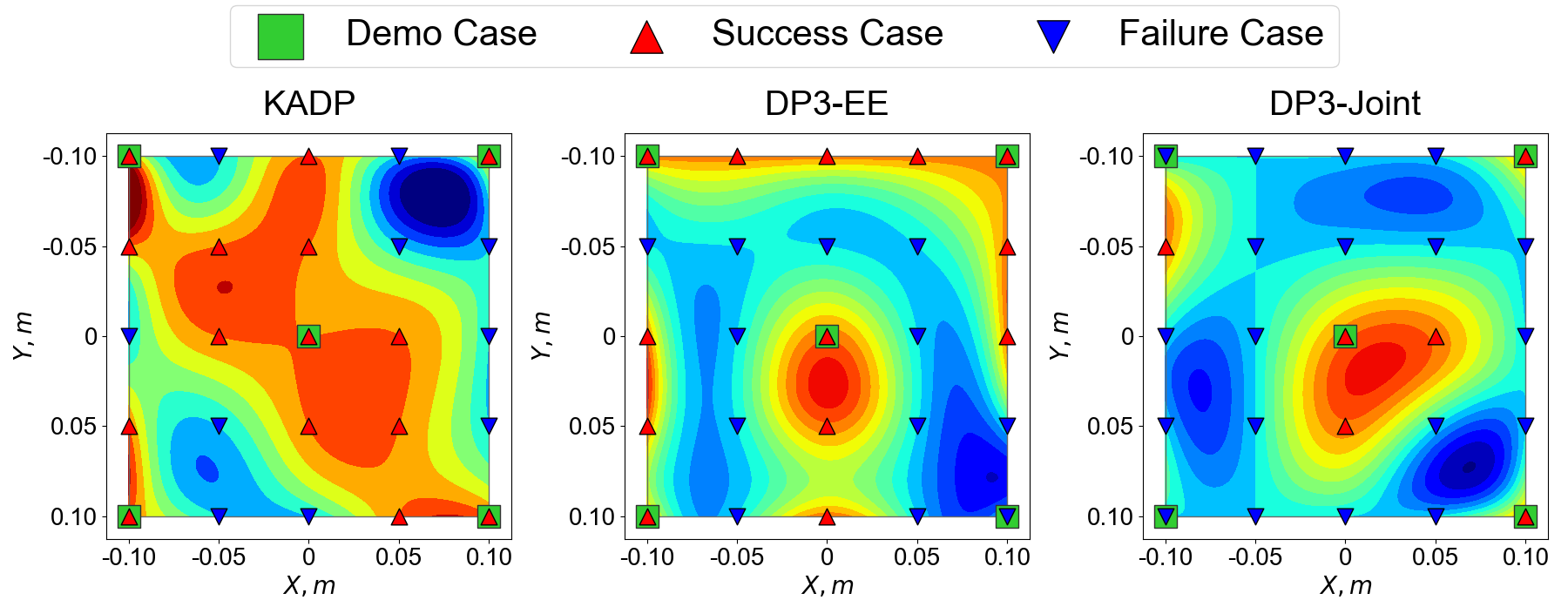}}
\hfill\\
\subfloat[Pick up Cube: 13 demonstrations]{\includegraphics[width=1.0\linewidth]{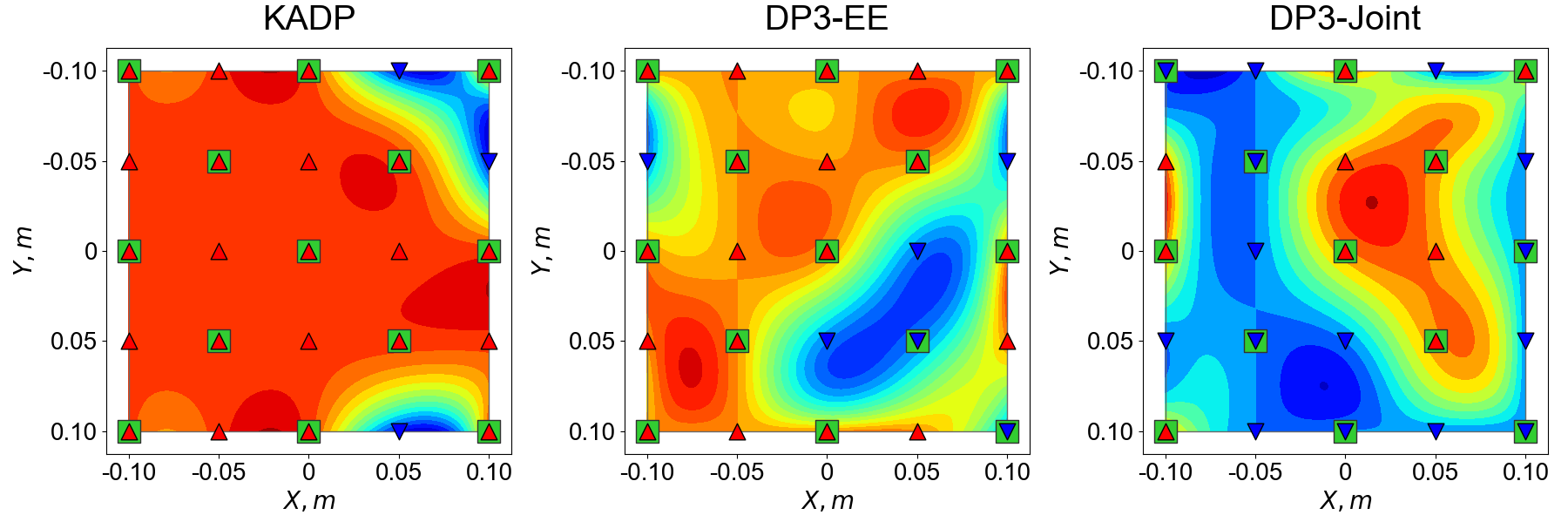}}
\caption{Spatial generalization performance on $\mathtt{pick \ up \ cube}$.}
\label{fig:heatmap}
\end{figure}

\textbf{Spatial Generalization and Sample Efficiency}:
In the $\mathtt{pick\; up\;cube}$ task, the initial positions of cube are constrained inside a 20cm $\times$ 20cm workspace, from which 25 positions are uniformly sampled for evaluation. Two demonstration settings are considered: 1) 5 demonstrations including the center and four corners of the workspace, and 2) 13 demonstrations including the center, four corners, four edge midpoints, and four midpoints between the center and corners. 
DP3-Joint yields the lowest performance under both settings as reported in Table \ref{tab:2}, confirming the difficulty of learning effective policy from joint space with limited data.
KADP achieves success rates of 60\% and 88\% for two settings, respectively, surpassing both two baselines. Generated via cubic interpolation over the discrete evaluation points, the heatmaps in Fig.\ref{fig:heatmap} visualize the spatial distribution of success (the red region) and failure (the blue region) evaluation cases. Notably, the success region of KADP is substantially larger than those of the baselines, clearly indicating its superior spatial generalizability within the demonstration coverage.

\begin{figure}[bt]
\centering
\subfloat[DP3-Joint]{\includegraphics[width=0.325\linewidth]{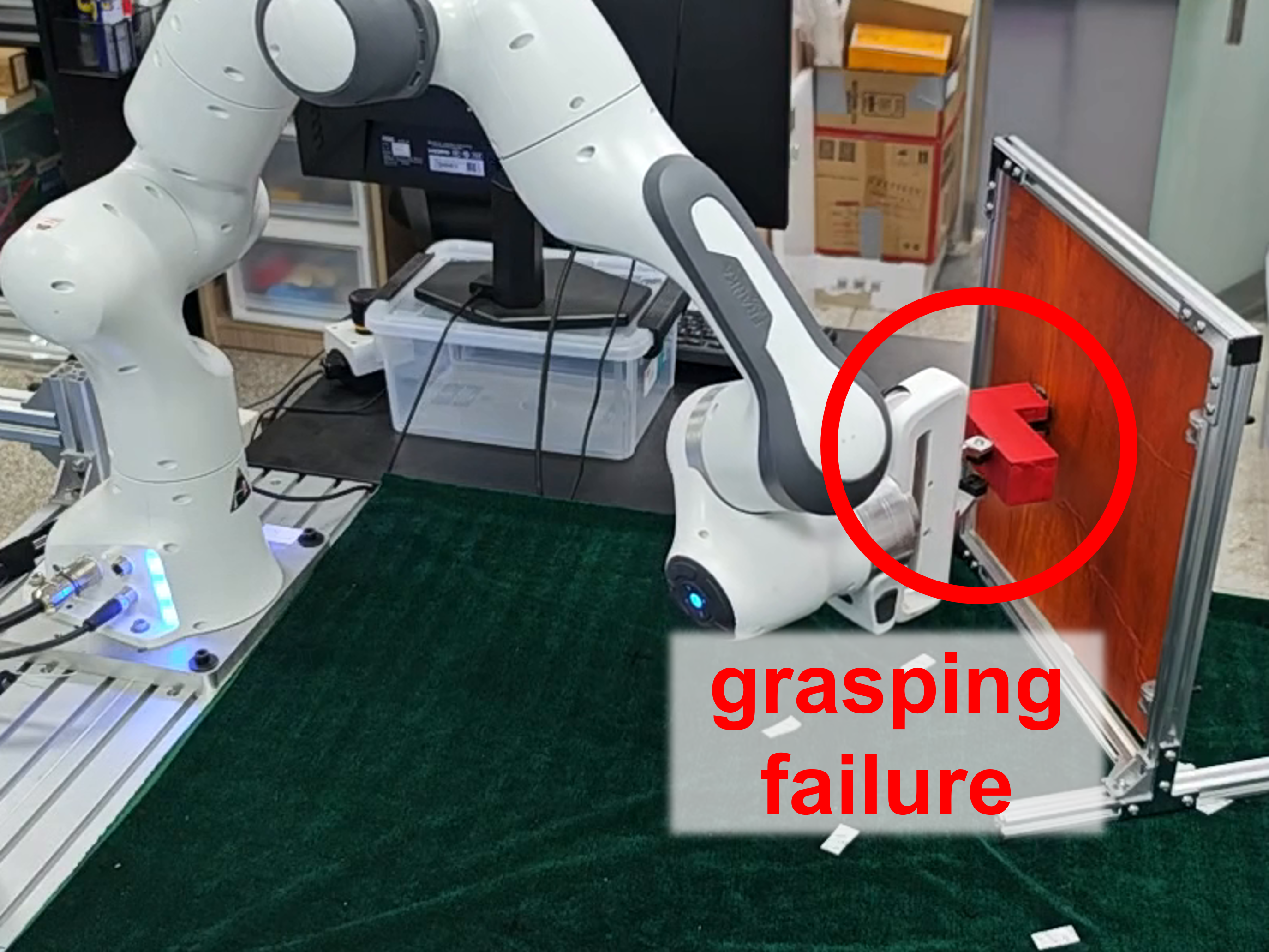}\label{fig:fail_open_door}}
\hfill
\subfloat[DP3-EE]{\includegraphics[width=0.325\linewidth]{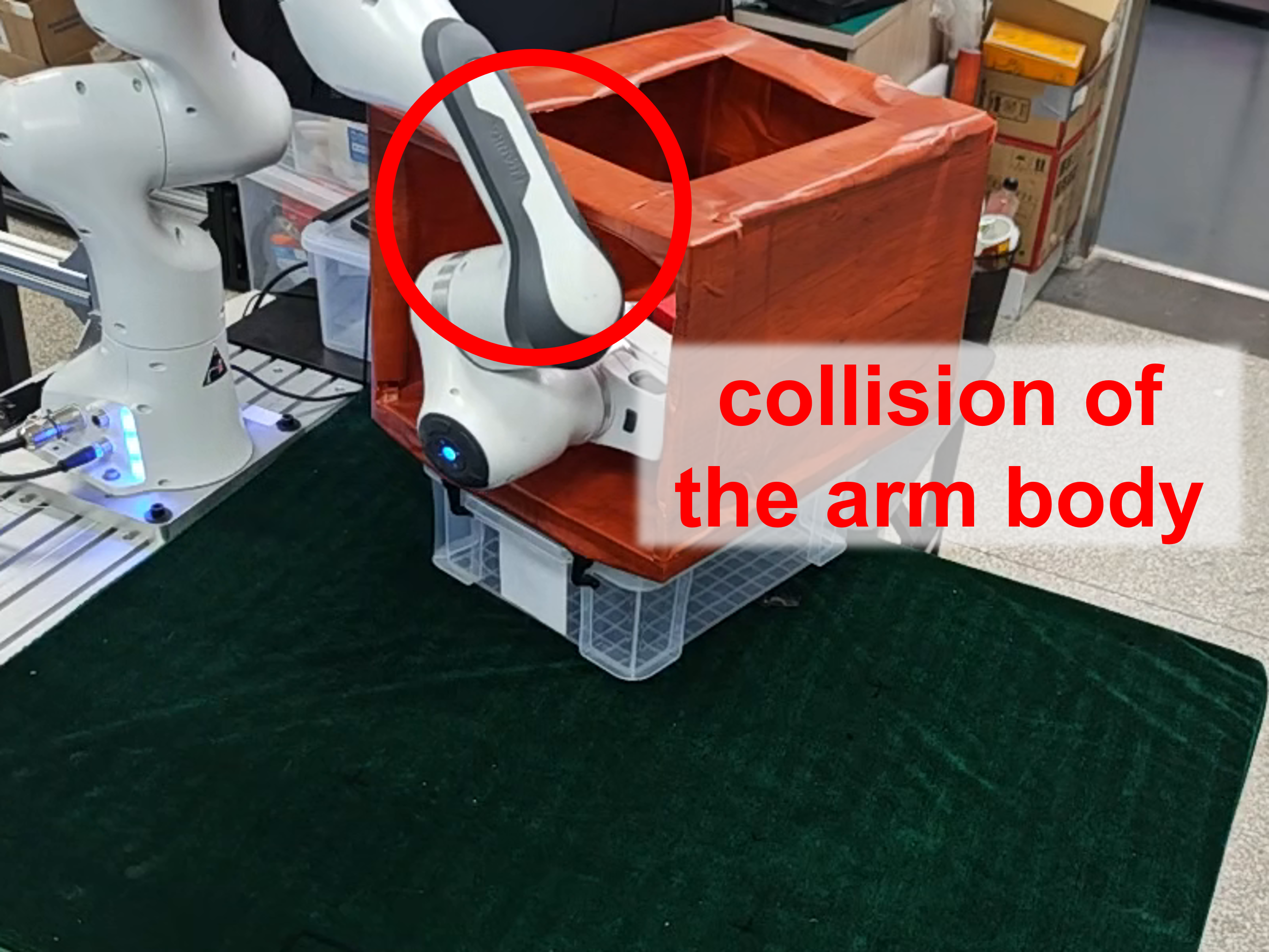}\label{fig:fail_cube_cabinet}}
\hfill
\subfloat[DP3-EE]{\includegraphics[width=0.325\linewidth]{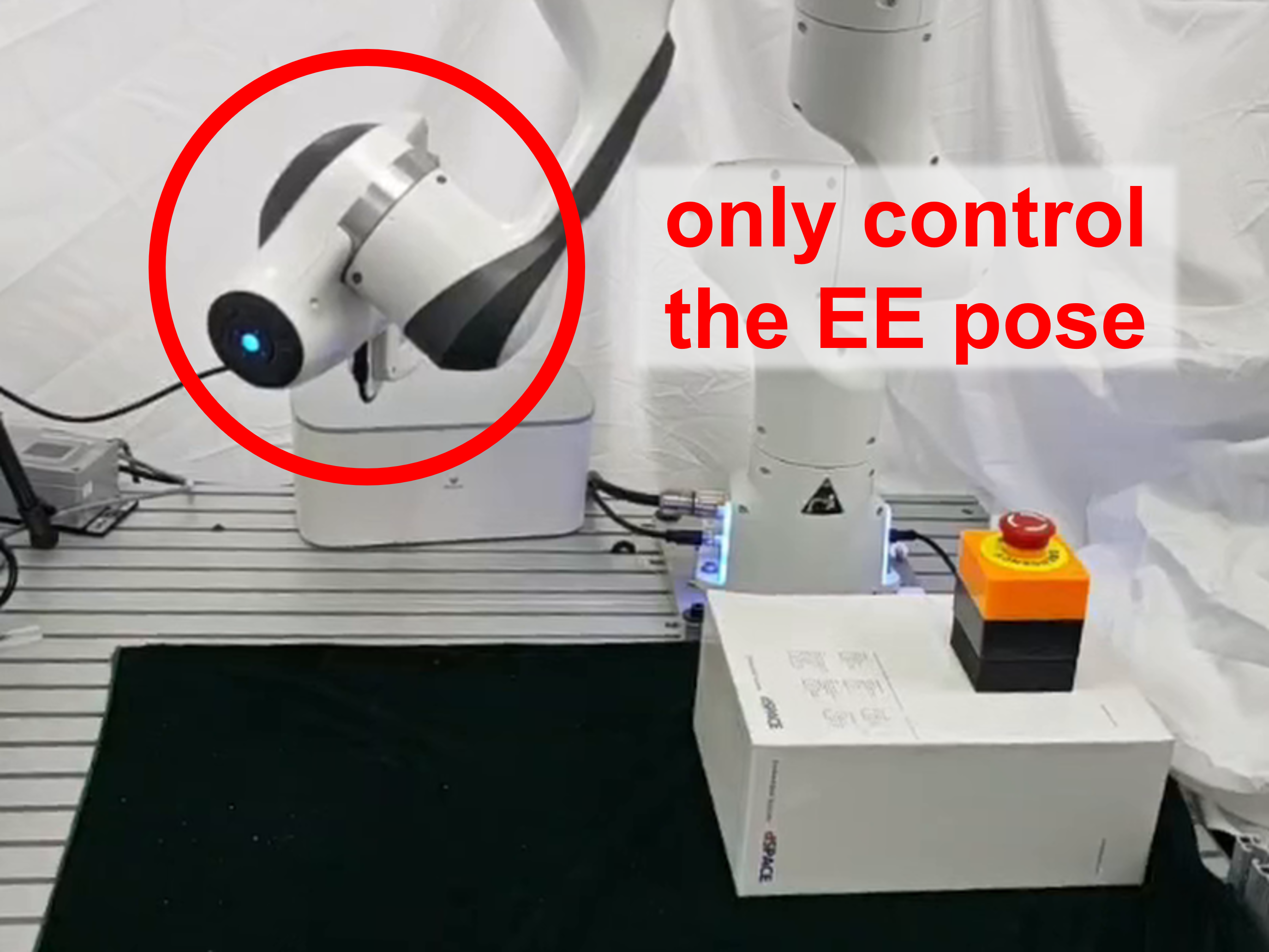}\label{fig:fail_push_button}}
\caption{Failure cases of baseline methods on the real-world tasks.}
\label{fig:fail}
\end{figure}

In the $\mathtt{open\;door}$ task, KADP and DP3-EE both successfully complete the task 8 times out of 10 trials, while DP3-Joint achieves a lower success rate of 60\%.
These results also suggest that KADP will not sacrifice the performance of end-effector-based policy when precisely controlling the gripper is sufficient, and offer better generalization ability and sample efficiency over joint-space learning. 
Fig.~\ref{fig:fail_open_door} illustrates a typical failure case of DP3-Joint, where inaccurate gripper position leads to unsuccessful grasping. In contrast, the failures from KADP and DP3-EE are attributed to difficult out-of-distribution door positions and orientations.

\textbf{Whole-Arm Manipulation}:
In the $\mathtt{put\;cube\;in\; cabinet}$ task, since the robot configuration cannot be fully controlled through end-effector poses alone, DP3-EE struggles on this task with only a 10\% success rate. As illustrated in Fig.\ref{fig:fail_cube_cabinet}, although the predicted end-effector pose is often suitable for insertion, frequent collisions with the top surface lead to task failures. DP3-Joint enables whole-body control, but the inaccurate task-space generalization often results in collisions with the cabinet’s side surfaces. In contrast, KADP effectively overcomes both challenges above, achieving a success rate of up to 90\%.

In the $\mathtt{push\; button\; elbow}$ task, DP3-EE fails in all trials, as it merely imitates end-effector trajectories without capturing the actual intent of the task.
As shown in Fig.\ref{fig:fail_push_button}, although the end-effector reaches a position similar to successful executions, DP3-EE is unable to control the elbow appropriately to manipulate the object. 
In contrast, both KADP and DP3-Joint achieve a 100\% success rate.
The comparable performance of DP3-Joint and KADP is expected, as the joint-space robot action is only 3-dimensional in this task, making the mapping from joint space to 3D space significantly easier to learn.


\section{Conclusion}

In this paper, we present Kinematics-Aware Diffusion Policy (KADP), an imitation learning framework that aligns task, observation, and action spaces in the consistent 3D space for effective whole-body robotic manipulation. 
By representing both robot states and actions as a set of 3D nodes on the robot arm, KADP improves sample efficiency and spatial generalization compared to end-effector-pose or joint-space approaches, while also enabling full-body control.
Extensive experiments in both simulation and real-world environments demonstrate the superiority of KADP in complex and body-aware manipulation tasks, underscoring its potential as a scalable and generalizable solution for learning whole-arm robot behaviors from limited demonstrations. 
Despite its effectiveness, KADP still has several limitations. 
Like all fixed-data imitation learning approaches, it is restricted to the distribution of the provided demonstrations and struggles with out-of-distribution generalization.
Additionally, while the node-based representation integrates well with the diffusion policy framework, its high dimensionality might limit compatibility with other policy learning paradigms such as reinforcement learning.
Future directions could involve exploring its performance in large-scale policy and extending it to long-horizon, multi-task imitation learning scenarios.


\bibliographystyle{IEEEtran}
\bibliography{ref.bib}

\end{document}